\documentclass[11pt]{article}

\usepackage[final]{acl}
\usepackage{tabularx}
\usepackage{booktabs}

\usepackage{times}
\usepackage{latexsym}

\usepackage[utf8]{inputenc}

\usepackage{microtype}

\usepackage{inconsolata}

\usepackage{graphicx}
\usepackage{multirow}
\usepackage{array}  

\usepackage[symbol]{footmisc}

\title{\Large{\textbf{L3Cube-MahaEmotions: A Marathi Emotion Recognition Dataset with Synthetic Annotations using CoTR prompting and Large Language Models}}}

\author{ Nidhi Kowtal\textsuperscript{1},
\textbf{Raviraj Joshi}\textsuperscript{2,3} \\
  \textsuperscript{1} Pune Institute of Computer Technology, Pune, Maharashtra India \\ 
  \textsuperscript{2} Indian Institute of Technology Madras, Chennai, Tamil Nadu India\\ 
  \textsuperscript{3} L3Cube Labs, Pune\\
   \texttt{\{kowtalnidhi, ravirajoshi\}@gmail.com}}

\begin{document}
\maketitle

\begin{abstract}
Emotion recognition in low-resource languages like Marathi remains challenging due to limited annotated data. We present L3Cube-MahaEmotions, a high-quality Marathi emotion recognition dataset with 11 fine-grained emotion labels. The training data is synthetically annotated using large language models (LLMs), while the validation and test sets are manually labeled to serve as a reliable gold-standard benchmark. Building on the MahaSent dataset, we apply the Chain-of-Translation (CoTR) prompting technique, where Marathi sentences are translated into English and emotion labeled via a single prompt. GPT-4 and Llama3-405B were evaluated, with GPT-4 selected for training data annotation due to superior label quality. We evaluate model performance using standard metrics and explore label aggregation strategies (e.g., Union, Intersection). While GPT-4 predictions outperform fine-tuned BERT models, BERT-based models trained on synthetic labels fail to surpass GPT-4. This highlights both the importance of high-quality human-labeled data and the inherent complexity of emotion recognition. An important finding of this work is that generic LLMs like GPT-4 and Llama3-405B generalize better than fine-tuned BERT for complex low-resource emotion recognition tasks. The dataset and model are shared publicly at \url{https://github.com/l3cube-pune/MarathiNLP}.
\end{abstract}

\section{Introduction}

Recent advances in NLP have mainly benefited high-resource languages like English and Chinese, which have ample data and annotations \cite{b1}. Low-resource languages, however, face challenges due to limited high-quality data and complex grammar, leading to poor model performance \cite{b2}. Even multilingual LLMs, effective in translation, struggle with direct prompts in these languages \cite{b3, b12}. We focus on Marathi, spoken by about 83 million people, which remains underrepresented in NLP due to scarce tools and datasets \cite{b13,b14,b15}. Its syntactic complexity adds to the modeling difficulty \cite{b3}.
\newline

To overcome these challenges, we created an emotion classification dataset for Marathi by leveraging the capabilities of large language models like GPT-4 and Llama3-405B. A key limitation in emotion classification for Marathi is the lack of labeled emotional datasets. Manual labeling is expensive and time-consuming, which makes progress in low-resource languages slower. To address this, we combined manual validation with annotation using LLMs to produce a high-quality dataset efficiently. We manually labeled the validation and test sets to ensure a gold-standard benchmark and also annotated these sets using GPT-4 and Llama3-405B to evaluate their performance. Since GPT produced more accurate results, we used it to annotate the training set as well, accelerating progress for Marathi NLP through the strategic use of LLMs.
\newline

Interestingly, we observe that GPT-4 significantly outperforms BERT-based models trained on its own generated labels. This indicates that fine-tuning smaller models on noisy or automatically annotated data does not necessarily lead to better performance than the original LLM. The results underscore the inherent complexity of multi-label emotion recognition—an intricate task where generic LLMs like GPT-4 are better equipped to capture subtle emotional cues than fine-tuned, smaller models. This contrasts with findings from \cite{b4}, where BERT models trained on clean, high-quality data were shown to outperform LLMs in low-resource scenarios. In our case, the presence of residual noise in the training labels or the complexity of the task itself likely hinders the ability of BERT-based models to generalize effectively.
\newline

\begin{figure*}[htbp]
  \centering
  \includegraphics[width=0.8\textwidth]{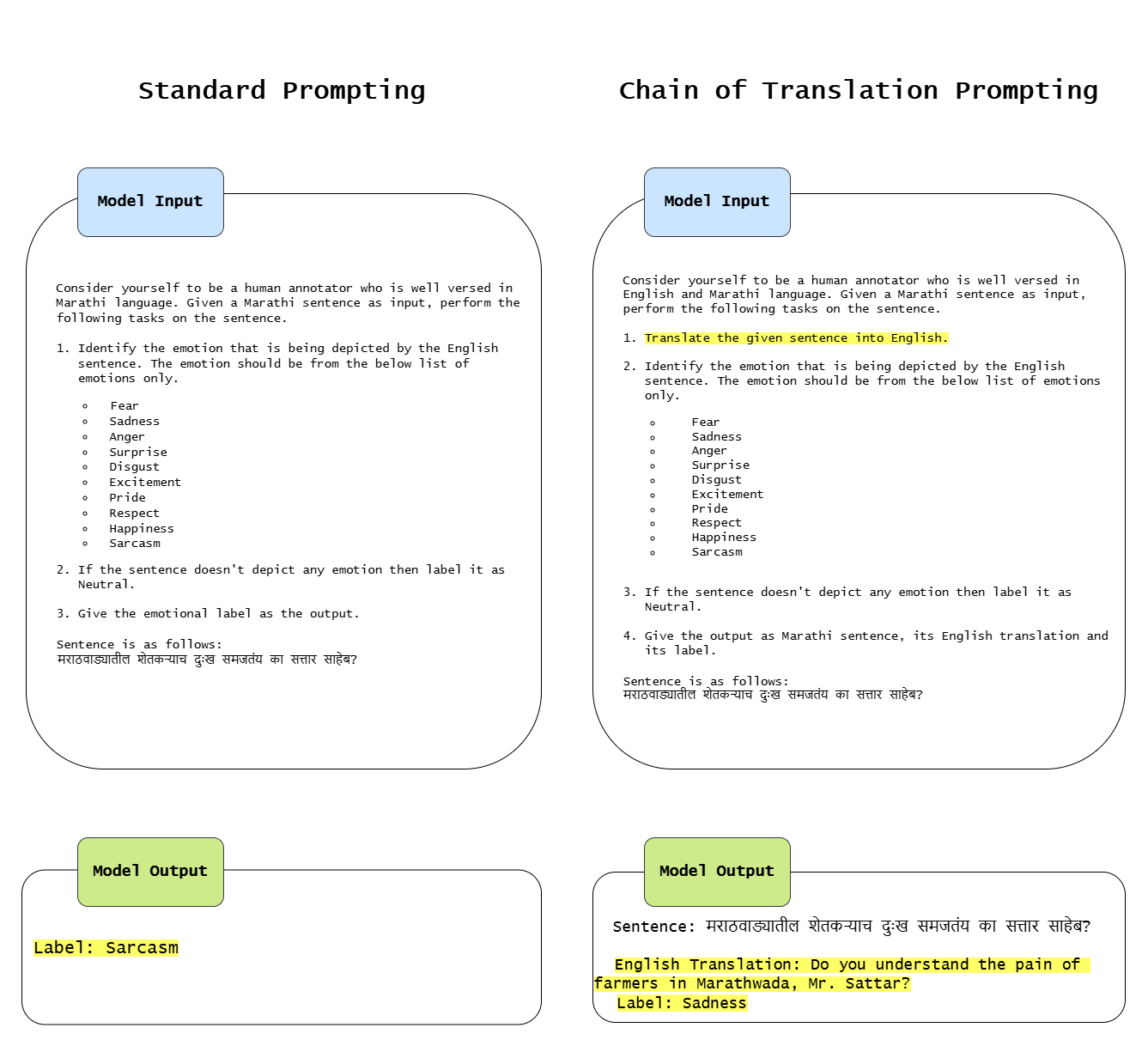}
  \caption{Chain of Translation Prompting (CoTR)}
  \label{fig:emotion_prompt}
\end{figure*}

We use a prompting technique called Chain-of-Translation Prompting (CoTR) to improve the quality of emotion annotation for a low-resource language like Marathi~\cite{b22}. The CoTR approach, illustrated in Figure~\ref{fig:emotion_prompt}, has been shown to outperform standard prompting strategies, and is adopted in this study for its effectiveness. Given the scarcity of Marathi training data, LLMs may struggle to accurately predict emotion labels directly from Marathi sentences. To address this, we translate Marathi inputs into English and then generate emotion labels using the translated text. This enables LLMs to leverage their stronger English language understanding. CoTR leads to more reliable emotion classification while preserving the intent of the original Marathi content. We independently validate its effectiveness on the MahaEmotions dataset.

The main contributions of this work are as follows:
\begin{itemize}
    \item We curate MahaEmotions\footnote{\url{https://github.com/l3cube-pune/MarathiNLP/tree/main/L3Cube-MahaEmotions}}\footnote{\url{https://huggingface.co/l3cube-pune/marathi-emotion-detect}}, a new Marathi Emotion Classification dataset\footnote{\url{https://huggingface.co/datasets/l3cube-pune/MahaEmotions}} annotated with eleven emotion categories, containing both model-generated and human annotated labels to ensure good annotation quality. The dataset consists of (12k, 1.5k, 1.5k) train, test, and validation samples, respectively.
    
    \item We use Chain-of-Translation (CoTR) as an effective prompting strategy to use multilingual LLMs for emotion tagging. Instead of direct categorization in Marathi, CoTR translates Marathi input into English before labeling the data, considerably enhancing the tagging accuracy. Notably, we observe an absolute 6\% improvement in the GPT-4 performance using CoTR prompting.

        \item We benchmark the performance of multiple models on this task, including GPT-4, LLaMA3-405B, and a fine-tuned MahaBERT-V2 model. Our results show that GPT-4 outperforms LLaMA-3, which in turn outperforms fine-tuned MahaBERT-V2, both in terms of accuracy and F1-score.

    \item We manually annotate a high-quality test set to evaluate how the LLMs perform on the tagging task.
\end{itemize}

\section {Related Work}
Low-resource languages have consistently faced challenges in NLP due to the lack of sufficient linguistic resources, standardized benchmarks, and annotated corpora. As a result, they remain significantly underrepresented in mainstream NLP research \cite{b18}.
The emergence of multilingual pretrained language models has helped address some of these issues through cross-lingual transfer, enabling better performance across languages with limited data.
\newline

Multilingual architectures such as mBERT, mT5, and XLM-R have shown reasonable zero-shot and few-shot performance on downstream tasks in low-resource settings \cite{b3, b20}. Prompt-based learning techniques have also proven effective in adapting pretrained models to new tasks, particularly in scenarios where task-specific fine-tuning is not feasible due to data scarcity \cite{b2}.
\newline

Recent advances like L3Cube-MahaNLP and MahaBERT have boosted research in syntactic parsing, classification, and sentiment analysis for Marathi by providing large monolingual datasets and transformer models \cite{b13, b15, pingle2023l3cube, kulkarni2021l3cubemahasent, velankar2022l3cube}. However, there’s still limited work on deeper tasks like emotion recognition. Marathi’s complex grammar and differences from English make it hard to directly apply models trained on high-resource languages. Similar trends appear in other low-resource languages like Khasi, where encoder-decoder transformers have improved tasks like translation despite limited data \cite{b1}.
\newline

In the broader NLP community, there has been growing interest in emotion recognition, especially in the context of multilingual and multimodal systems. However, similar advancements for Marathi are still quite limited. In comparison, significant progress has been made for Hindi and Hindi-English code-mixed text, with several emotion classification models and datasets available \cite{b8, b9, b10}. A good example is the EmoInHindi corpus, a low-resource benchmark that provides multi-label emotion annotations along with dialogue-level context \cite{b10}. Recent surveys also highlight the importance of using customized model architectures, cross-lingual transfer, and domain adaptation techniques for improving emotion classification in low- and mid-resource languages \cite{b21}.
\newline

Additionally, research on LLMs’ reliability for non-English inputs is ongoing. Zhang et al. \cite{b12} critically assess GPT-4 and other LLMs, showing performance drops for underrepresented, morphologically rich languages like Marathi. This questions whether such models can be directly used for low-resource emotion recognition without translation or augmentation. Prompt engineering adapted to linguistic traits \cite{b17} offers a practical way to overcome these limits. In multilingual contexts, translation-based prompting notably improves semantic understanding and emotion consistency.
\newline

In this study, we expand on these discoveries and provide a Chain-of-Translation (CoTR) prompting architecture for Marathi text emotion recognition that makes use of multilingual language models \cite{b22}. Our method uses a single prompt that first translates the Marathi sentence into English and then predicts the emotion using English-based prompt templates.

\section{Methodology}

\begin{figure*}[htbp]
  \centering
  \includegraphics[width=0.8\textwidth]{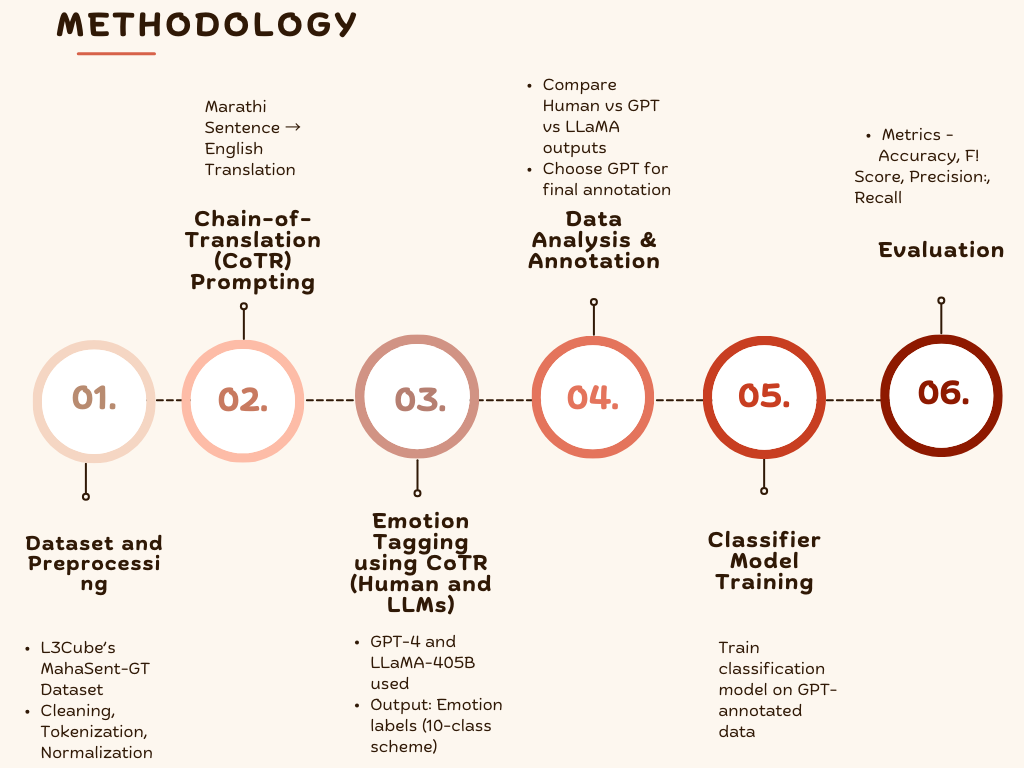}
  \caption{Emotion Tagging using Human and LLMs (CoTR)}
  \label{fig:flowchart}
\end{figure*}

Our methodology involves curating a high-quality Marathi emotion dataset, applying Chain-of-Translation (CoTR) prompting for emotion tagging using both human annotators and multilingual LLMs, and training a classifier on the annotated data. As shown in Figure~\ref{fig:flowchart}, the process includes dataset preprocessing, CoTR-based emotion labeling, model comparisons, and final evaluation using standard classification metrics.

\subsection{Dataset Description}

For this study, we have used the publicly available L3Cube's MahaSent-GT dataset \cite{b13}, a sentiment analysis corpus in Marathi. The dataset contains textual content primarily sourced from Twitter. Each sentence is originally labeled with sentiment (Positive, Negative, Neutral), and we extend this dataset by introducing emotion labels.
The dataset contains a total of 15,000 Marathi sentences. It provides a suitable foundation for emotion classification tasks due to its coverage of real-world, emotion-rich textual inputs. The distribution of emotion labels across the train, validation, and test sets, along with example sentences, is shown in Table~\ref{tab:emotion_distribution}.

\subsection{Emotion Label Taxonomy and Annotation Scheme}

We utilize a fixed set of eleven basic emotion labels: \textit{Happiness, Sadness, Anger, Fear, Surprise, Disgust, Excitement, Pride, Respect, Sarcasm}, and \textit{Neutral}. A careful selection process was used to ensure that this set of emotions was both simple enough to allow for consistent classification over a large number of phrases and expressive enough to reflect a wide range of sentiments.
\newline

A label is assigned to each sentence in the dataset according to the primary emotion it conveys. Although this set of emotion labels is based on popular psychological models like Ekman’s basic emotions and Plutchik’s emotion wheel, it is a simplified version made for practical use. Marathi is a diverse language, with many emotional states that are hard to define in a pre-defined set of categories. 
\newline

Emotions such as \textit{Trust}, \textit{Anticipation/Hope}, \textit{Contempt}, \textit{Curiosity}, \textit{Inspiration}, \textit{Disappointment}, \textit{Deep yearning}, \textit{Gentle sorrow}, \textit{Emotional overwhelm}, \textit{Helplessness}, \textit{Separation-induced longing}, \textit{Contentment}, \textit{Melancholy}, and \textit{Deep inner experience} are a few complex emotions that are subtle, context-dependent, making them difficult to represent in standard NLP frameworks. Although these could theoretically be classified as distinct emotion categories, we chose to concentrate on a more manageable and useful set of labels. After carefully reviewing the dataset and performing manual analysis and preprocessing, we selected emotion categories that were most frequently observed in day-to-day usage and could be annotated consistently at scale.
\newline

Since there are not many extensive emotion datasets for Marathi, we used this set of eleven emotions to enable consistent and scalable annotation. Both language models and human annotators benefit from this fixed list since it helps them concentrate on distinct, non-overlapping categories. The primary emotion that each sentence in the sample conveys is labeled. The strongest or most obvious emotion is selected when a text comprises multiple emotions. 

\subsection{Prompt-Based Annotation Strategy}

We designed a structured prompt to guide the language model during tagging. Since most large language models (LLMs) are trained primarily on English, we include translation in the same prompt. The Marathi sentence is first translated to English, and then the model predicts the emotion from a predefined set of categories: \textit{Fear, Sadness, Anger, Surprise, Disgust, Excitement, Pride, Respect, Happiness, Sarcasm}, and \textit{Neutral}. If a sentence contains more than one emotion, the most prominent one is assigned to it. If no emotion is clearly expressed, the sentence is labeled as \textit{Neutral}.

\subsection{Models Used}

\begin{enumerate}
    \item \textbf{GPT-4o:}
    \newline
     GPT-4o is developed by OpenAI, with ~1.8 trillion parameters (unofficial). It is a closed-source model and accessible through APIs provided by OpenAI. GPT-4o builds on the advancements of its previous versions, offering enhanced capabilities in natural language understanding, generation, and reasoning across a wide range of tasks.
    \item \textbf{Llama 3.1 405B:}
    \newline
    Llama 3.1 (Large Language Model for Multilingual Applications) is the third iteration in the Meta Llama series, designed with multiple variants, including a 405 billion parameter version and an 8 billion parameter version. These models are typically open-source. Llama3 models are optimized for multilingual tasks, incorporating vast and diverse datasets to improve performance across different languages.
    \item \textbf{MahaBERT-V2:}
    \newline
    MahaBERT-V2 is a transformer-based language model pre-trained specifically on a large corpus of Marathi text. It captures rich morphological and syntactic patterns of the Marathi language, making it well-suited for downstream NLP tasks in Marathi. Despite being domain-specific, its performance on emotion classification was moderate, with an accuracy of 63\% and an F1 score of 0.47.

     \item \textbf{MuRIL:}
    \newline
   MuRIL (Multilingual Representations for Indian Languages) is a multilingual BERT model developed by Google, trained on 17 Indian languages including Marathi. It supports both transliterated and native scripts, and enables zero-shot and multilingual transfer learning. In our experiments, MuRIL achieved an accuracy of 60\% and an F1 score of 0.42, slightly underperforming compared to MahaBERT-V2, likely due to its generalization across many languages rather than specialization in Marathi.

\end{enumerate}

 \begin{table*}[h]
\centering
\includegraphics[width=1\textwidth]{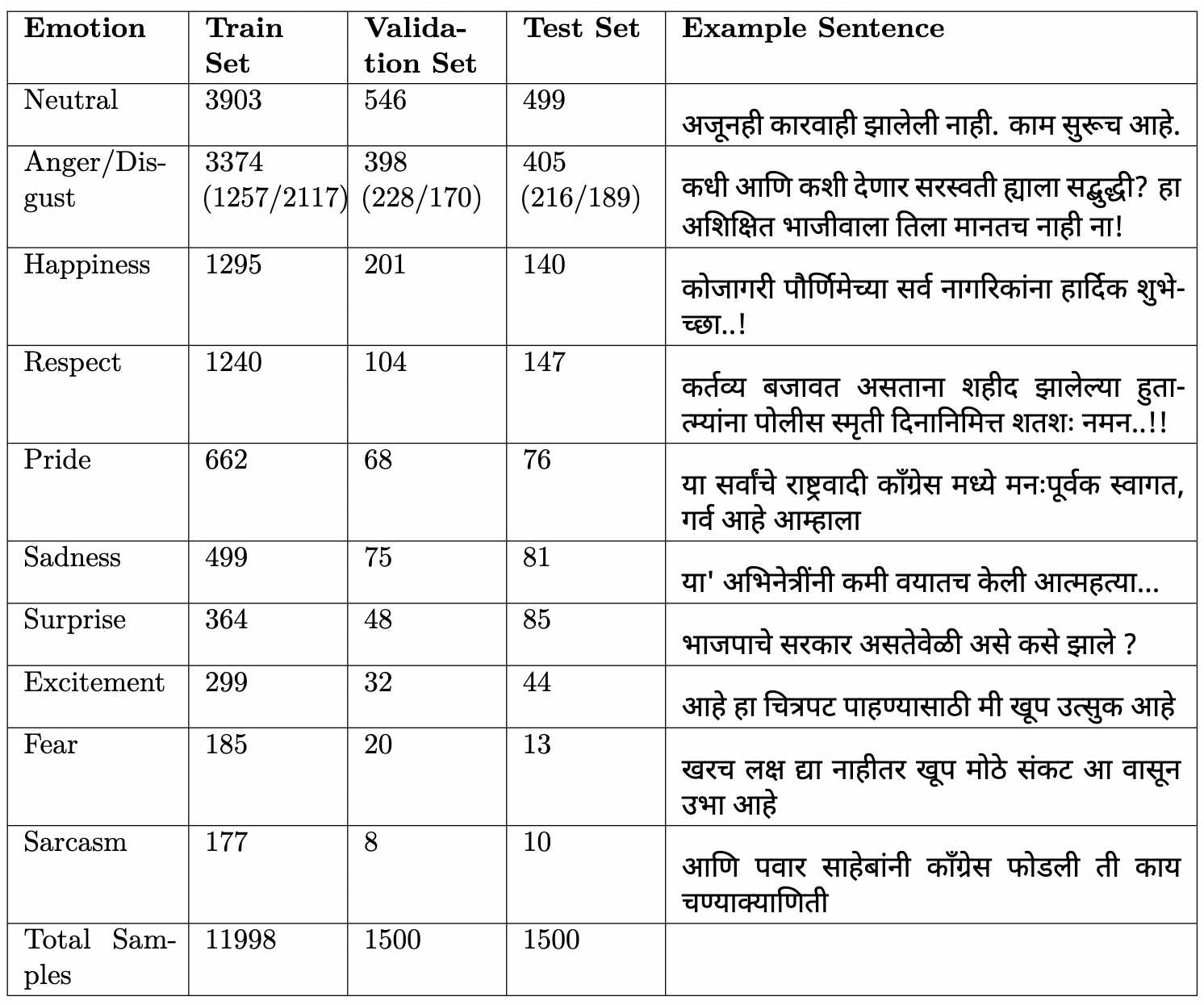}
\caption{Number of samples per emotion label in the train, validation, and test sets, along with example sentences}
  \label{tab:emotion_distribution}
\end{table*}

\begin{table}[h!]
\centering
\begin{tabularx}{\linewidth}{|p{3cm}|X|X|}
\hline
\textbf{Statistics} & \textbf{Validation Set} & \textbf{Test Set} \\ \hline
GPT-4 Correct & 1265 & 1284 \\ \hline
Llama Correct & 962 & 1051 \\ \hline
Llama Correct, GPT-4 Incorrect & 106 & 100 \\ \hline
GPT-4 Correct, Llama Incorrect & 409 & 333 \\ \hline
Both Correct & 856 & 951 \\ \hline
At Least One Correct & 1371 & 1384 \\ \hline
Both Incorrect & 129 & 116 \\ \hline
\end{tabularx}
\caption{Model performance statistics for validation and test sets (each containing 1500 sentences)}
\label{tab:model_stats}
\end{table}

\begin{table*}[h!]
\centering
\begin{tabularx}{\linewidth}{|X|X|X|X|X|}
\toprule
\textbf{Model} & \textbf{Accuracy} & \textbf{Precision} & \textbf{Recall} & \textbf{F1 Score} \\ \midrule
MahaBERT-V2 & 0.63 & 0.65 & 0.62 & 0.64 \\ \hline
MuRIL & 0.59 & 0.62 & 0.59 & 0.60 \\ \hline
GPT-4 & 0.83 & 0.85 & 0.82 & 0.83 \\ \hline
GPT-4 (CoTR) & \textbf{0.86} & \textbf{0.88} & \textbf{0.85} & \textbf{0.86} \\ \hline
Llama3-405B (CoTR) & 0.70 & 0.74 & 0.70 & 0.72 \\ \bottomrule
\end{tabularx}
\caption{Evaluation metrics for different models on the MahaEmotions test set (Weighted metrics). Note that Anger and Disgust are merged into a single class during both training and evaluation.}
\label{tab:model_metrics}
\end{table*}

\begin{figure*}[htbp]
  \centering
  \includegraphics[width=0.5\textwidth]{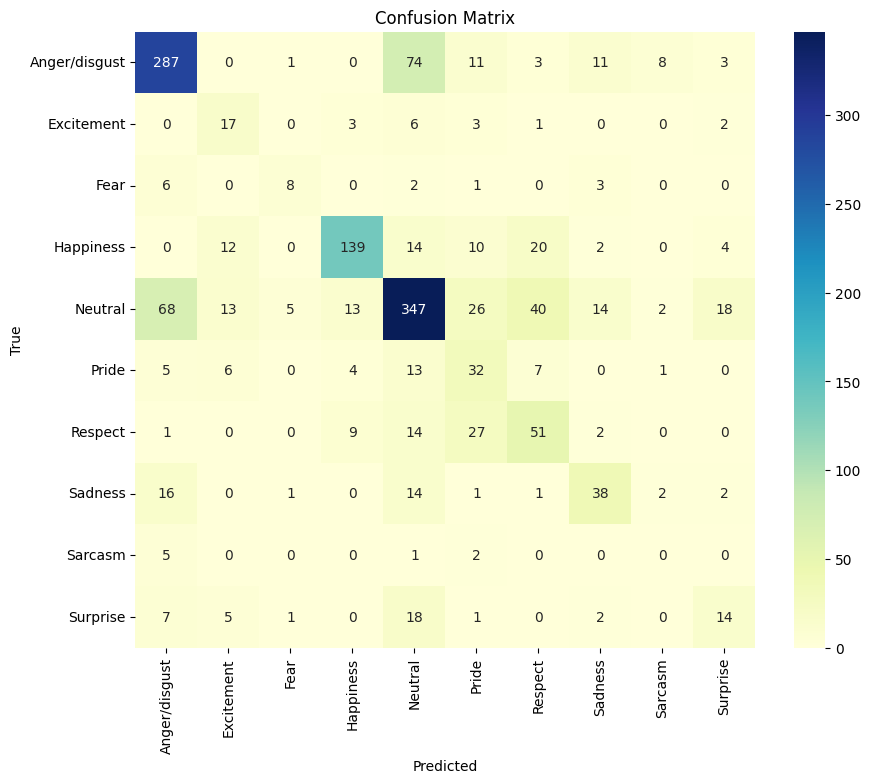}
  \caption{Confusion matrix for MahaEmotions classification task using L3Cube's MahaBERT-V2}
  \label{fig:conf_matrix}
\end{figure*}
\section{Results}

\subsection{Gold Test Set}
We manually annotated test and validation sets containing 1500 sentences each. These human annotations are treated as the ground truth for evaluating model performance.

\subsection{GPT-4 vs Llama3-405B}
We evaluated the performance of GPT-4 and Llama-405B by prompting each model individually to classify the same set of sentences. The classification was performed after translating the Marathi inputs into English. A model’s prediction was considered correct only if it matched the human-provided label.

We considered multiple evaluation scenarios:
\begin{itemize}
    \item \textbf{Correct Prediction:} The model label matches the human-annotated gold label.
    \item \textbf{Disagreement Resolution:} If both models gave labels different from the human label, no credit was given to either.
    \item \textbf{Overlap Analysis:} We analyzed agreement and disagreement patterns, including cases where both models were correct, only one was correct, or both were incorrect.
\end{itemize}

Based on the comparative analysis of these models, we found GPT-4 to be the more consistent and accurate model. Since the performance of GPT-4 alone was comparable to the combination of GPT-4 and Llama3-405B, we chose GPT-4 for the large-scale annotation of the training dataset.

We evaluated the performance of GPT-4 and Llama3-405B on both the validation and test datasets, each consisting of 1500 Marathi sentences. Table~\ref{tab:model_stats} summarizes the correctness statistics across both models.

GPT-4 consistently outperformed Llama in both validation and test set. On the test set, GPT-4 correctly classified 1284 sentences, while Llama correctly classified 1051. GPT-4 showed better accuracy, with 333 instances where GPT-4 was correct and Llama was incorrect, compared to only 100 instances where Llama was correct and GPT-4 was wrong.

Given the OR of both models' predictions is similar with GPT-4’s performance (1384 for OR vs. 1284 for GPT), we decided to tag the training data exclusively using GPT-4 for the final classifier model.

After annotation, we trained a classifier on the GPT-labeled dataset. The overall performance of this classifier on the test set is shown in Table~\ref{tab:model_metrics}. It achieved an accuracy of 63\%, with a precision of 0.65, recall of 0.62, and F1 score of 0.64. The detailed confusion matrix is presented in Figure~\ref{fig:conf_matrix}, which shows the classification behavior across all emotion categories

The confusion matrix shows that all emotion categories are sometimes predicted as \textit{Neutral}. This is expected, as many Marathi sentences have emotions that are expressed in a very subtle way, making them harder for the model to detect. In such cases, the model often chooses the Neutral label. This problem is common in low-resource languages, where emotions depend more on cultural and contextual clues than on clear emotional words.

There are also some clear patterns of confusion between emotions that are similar in meaning. For example, \textit{Pride} and \textit{Respect} are often mixed up. In Marathi, pride is often expressed with respectful language, and respectful statements can sound like pride. Similarly, \textit{Happiness} and \textit{Excitement} are confused with each other because they are both positive emotions, with the main difference being the level of intensity. \textit{Fear} and \textit{Surprise} are also mixed up, as both can be caused by unexpected events.

In some cases, \textit{Sarcasm} is classified as \textit{Anger/Disgust}, which makes sense because sarcasm can carry a tone of irritation or contempt. \textit{Sadness} is sometimes labeled as \textit{Neutral} when expressed in a mild way, and as \textit{Anger/Disgust} when it includes frustration. These patterns show two main challenges: the tendency of the \textit{Neutral} class to attract unclear cases, and the difficulty of separating emotions that are similar in meaning or context. Better use of context and targeted data augmentation could help improve the model’s performance in these cases.

\subsection{Chain of Translation Prompting (CoTR) vs Non-CoTR Approach}
As shown in Table~\ref{tab:model_metrics}, using CoTR leads to consistent improvements in accuracy, precision, recall, and F1 score. By translating Marathi inputs into English, multilingual LLMs can more effectively apply their English-language capabilities, enhancing emotion classification performance in low-resource languages like Marathi.

\section{Limitations}
One limitation of our work is that we used large language models (LLMs) that are mostly trained on English or multilingual data, not specifically on Marathi. Because of this, the models may not fully understand the deeper meanings or cultural context in Marathi sentences.

Another limitation is that our dataset has fewer examples of rare or complex emotions, which makes it harder for the model to learn and predict such emotions correctly. Emotions like Emotional overwhelm and Gentle sorrow are especially difficult to label consistently.

\section {Future Work and Conclusion}
In this work, we focused on the task of emotion classification for Marathi, a low-resource language. We created a high-quality dataset by combining predictions from large language models (LLMs) like GPT-4 and Llama-405B with manual checks. To improve accuracy, we used a method called Chain-of-Translation (CoTR), where Marathi sentences were first translated to English before labeling. GPT-4 showed consistent and reliable results, which made it suitable for large-scale annotation.
\newline

In the future, we plan to train LLMs using more Marathi-specific emotion data. This will help the models better understand the language and its emotional tone. We also want to include more sentences that show complex and subtle emotions of Marathi.

We also aim to test our Chain-of-Translation (CoTR) method on more LLMs such as Gemma, Grok, DeepSeek, and Mistral, to see how well it works with other models.
\section*{Acknowledgments}

This work was done under the mentorship of Mr. Raviraj Joshi (Mentor, L3Cube Pune). I would like to express our gratitude towards him for his continuous support and encouragement.

\bibliography{main}
\nocite{*}

\end{document}